\documentclass[10pt,twocolumn,letterpaper]{article}

\usepackage[pagenumbers]{cvpr} 

\definecolor{cvprblue}{rgb}{0.21,0.49,0.74}
\usepackage[pagebackref,breaklinks,colorlinks,allcolors=cvprblue]{hyperref}

\usepackage{dsfont}
\usepackage{url}
\usepackage[utf8]{inputenc} 
\usepackage[T1]{fontenc}    
\usepackage{hyperref}       
\usepackage{url}            
\usepackage{booktabs}       
\usepackage{amsfonts}       
\usepackage{nicefrac}       
\usepackage{microtype}      
\usepackage{xcolor}         

\usepackage{graphicx}
\usepackage{amsmath}
\usepackage{amssymb}
\usepackage{booktabs}

\usepackage{array}
\usepackage{caption}
\usepackage{multirow}
\usepackage{makecell}
\usepackage{pifont}

\usepackage{colortbl}
\usepackage{enumitem}
\usepackage[normalem]{ulem}
\usepackage{adjustbox}
\usepackage{cellspace}
\usepackage{float}
\usepackage{amssymb}
\usepackage{booktabs}
\usepackage{cvpr}
\usepackage{wrapfig}
\usepackage{algorithm}
\usepackage{algorithmic}
\usepackage{eqparbox}

\def\paperID{7410} 
\def\confName{CVPR}
\def\confYear{2025}

\newlength\savewidth

\makeatletter
\g@addto@macro\@maketitle{
    \begin{figure}[H]
        \setlength{\linewidth}{\textwidth}
        \setlength{\hsize}{\textwidth}
        \centering

        \vspace{-10mm}

        \includegraphics[width=\textwidth]{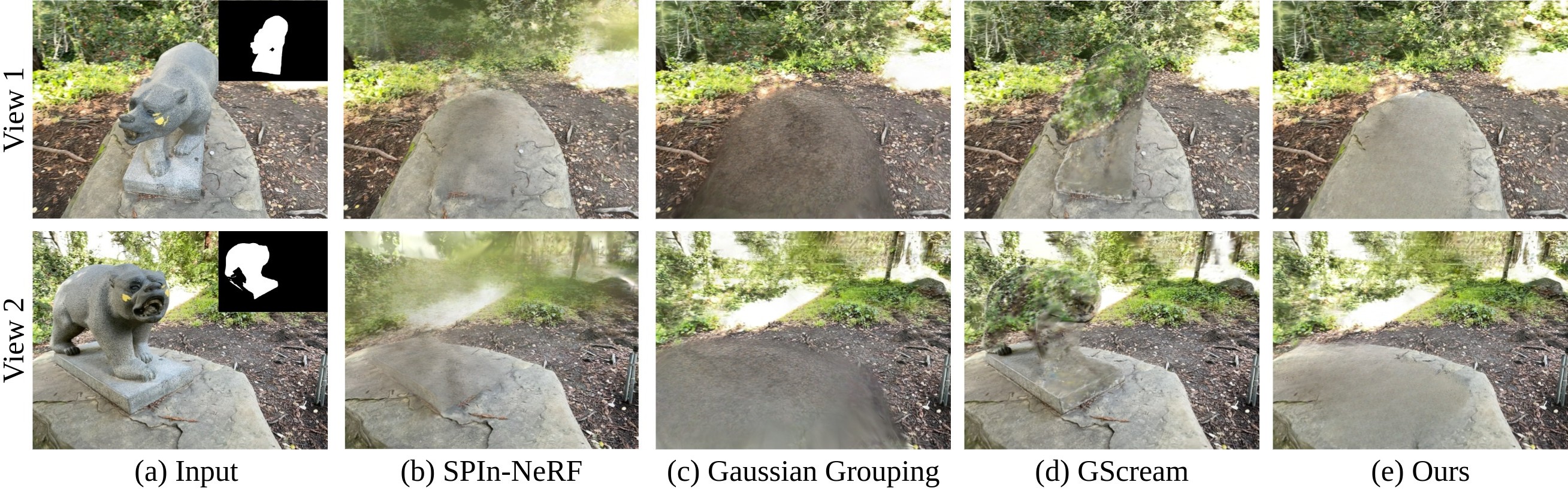}
        \vspace{-6mm}
        \caption{Given multi-view images of a scene and the object masks describing the object to be removed, our 3D Gaussian Inpainting with Depth-Guided Cross-View Consistency produces high-fidelity cross-view inpainting results. Compared with current state-of-the-arts such as SPIn-NeRF~\cite{mirzaei2023spin}, Gaussian Grouping~\cite{ye2023gaussiangrouping}, and GScream~\cite{wang2024gscream}, inpainting results of our method not only preserve visible background contents but also exhibit satisfactory consistency across camera views.}
        \label{fig:teaser}
    \end{figure}
}
\makeatother

\title{3D Gaussian Inpainting with Depth-Guided Cross-View Consistency}

\author{Sheng-Yu Huang$^{1,\dagger}$,\quad Zi-Ting Chou$^{2}$,\quad Yu-Chiang Frank Wang$^{1, 2,\ddagger}$, \\
\normalsize \textsuperscript{1} Graduate Institute of Communication Engineering, National Taiwan University \quad \textsuperscript{2} NVIDIA, Taiwan\\
{\tt\small $^{\dagger}$f08942095@ntu.edu.tw  $^{\ddagger}$frankwang@nvidia.com}
}

\begin{document}
\maketitle
\begin{abstract}
When performing 3D inpainting using novel-view rendering methods like Neural Radiance Field (NeRF) or 3D Gaussian Splatting (3DGS), how to achieve texture and geometry consistency across camera views has been a challenge. In this paper, we propose a framework of 3D Gaussian Inpainting with Depth-Guided Cross-View Consistency (3DGIC) for cross-view consistent 3D inpainting. Guided by the rendered depth information from each training view, our 3DGIC exploits background pixels visible across different views for updating the inpainting mask, allowing us to refine the 3DGS for inpainting purposes. 
Through extensive experiments on benchmark datasets, we confirm that our 3DGIC outperforms current state-of-the-art 3D inpainting methods quantitatively and qualitatively.   

\end{abstract}    
\section{Introduction}
\label{sec:intro}

Novel view synthesis for 3D scenes plays a vital role in 3D reconstruction and scene understanding. Recent advancements, such as Neural Radiance Fields (NeRF)~\cite{mildenhall2021nerforiginal, barron2021mip, yu2021plenoctrees, muller2022instant, sun2022direct, fridovich2022plenoxels, martin2021nerf, reiser2021kilonerf, chen2022tensorf} and 3D Gaussian Splatting (3DGS)~\cite{yu2024mipsplat, qin2024langsplat, ye2023gaussiangrouping, kerbl202333dgs, chen2024surveygs}, enable high-fidelity novel views by modeling volumetric properties. However, practical VR/AR applications~\cite{macedo2021ar1, broll2022augmentedar2} require more than reconstruction: they need editing capabilities that these methods do not fully address. Among editing challenges, object removal and inpainting~\cite{wang2024gscream, lin2024maldnerf} are particularly difficult, as direct removal creates visible holes, compromising visual quality. While 2D inpainting~\cite{lama, corneanu2024latentpaint, lugmayr2022repaint, podell2023sdxl, xie2023smartbrush, yang2023paintbyexample, ldm, yang2023unipaint} across multiple views is possible, maintaining consistency remains problematic, leading to artifacts and reduced fidelity. Thus, achieving seamless, multi-view-consistent inpainting for 3D scenes is still an open challenge.

As a pioneering work in 3D scene inpainting, SPIn-NeRF~\cite{mirzaei2023spin} proposes to use a pre-trained segmentation network~\cite{hao2021edgeflow} to generate plausible 2D inpaint masks for multi-view images with sparse human annotations of the object to be removed. However, as noted in subsequent research~\cite {lin2024maldnerf, wang2024gscream}, SPIn-NeRF and similar approaches~\cite{weder2023removingnerf, yin2023ornerf} rely heavily on 2D inpainting of multiple views separately, which hinders the cross-view consistency of the 3D inpainting results. To ensure cross-view consistency, RefNeRF~\cite{mirzaei2023referenceinpaint}
projects the inpainted image from a specific reference view onto other views using depth-guided projection, thereby ensuring more consistent inpainting results across views. Despite these advancements, these methods still require human-annotated 2D masks or sparse annotations to delineate the objects to be removed and the regions to be inpainted, making the process labor-intensive and limiting the scalability and practicality of these techniques.

To reduce the need for human annotation for obtaining inpainting masks, recent methods~\cite{ye2023gaussiangrouping, yin2023ornerf} tend to leverage the Segment Anything Model (SAM)~\cite{kirillov2023sam} 
models with NeRF or 3DGS to obtain 2D inpainting masks for multi-view images directly. Although these methods ease the requirement of human annotations for inpainting masks, they still rely on 2D inpainting results for different views as supervision, limiting the multi-view consistency of the inpainted 3D representations. To alleviate this limitation, some approaches~\cite{lin2024maldnerf, chen2024mvip, wang2024gscream, mirzaei2024reffusion, liu2024infusion, weber2024nerfiller} attempt to build a cross-view consistent 3D inpainting method on top of the 2D inpainting mask obtained from SAM. By either leveraging 2D diffusion models as perceptual guidance for the inpainted region~\cite{chen2024mvip, lin2024maldnerf, weber2024nerfiller} or ensuring feature consistency of corresponding pixels across different views~\cite{wang2024gscream}, these methods are able to produce more consistent 3D inpainting results without the requirement of human-annotated 2D inpainting masks. Nevertheless, most of the aforementioned methods rely on the provided per-scene 2D inpainting masks (either from human annotation or from SAM) for each view, which can include areas visible in other views, as mentioned in~\cite{ye2023gaussiangrouping}. As a result, the inpainted content within this area might be inconsistent across camera views, producing artifacts in the reconstructed 3D scene. 

In this paper, we propose a 3D Gaussian Inpainting with Depth-Guided Cross-View Consistency (3DGIC) to optimize the 3DGS model while achieving multi-view consistent and high-fidelity 3D inpainting with depth-guided inpainting masks to locate the inpainting region. Given a set of images of a scene with corresponding camera views and the object masks indicating an unwanted object in the scene (obtained from SAM~\cite{kirillov2023sam}, for example), our 3DGIC conducts the process of \textit{Inferring Depth-Guided Inpainting Masks} to consider depth information from all training views and refine the inpainting mask by discovering background pixels from different views. The refined inpainting masks are then used to provide a joint update of inpainting results and the underlying 3DGS model via \textit{3D Inpainting with Cross-View Consistency}. 
Through experiments on real-world datasets, we quantitatively and qualitatively demonstrate that our 3DGIC performs favorably against state-of-the-art NeRF/3DGS-based inpainting methods by achieving better fidelity and multi-view consistency.

The key contributions of our approach are as follows:
\begin{itemize}
\item We propose a 3D Gaussian Inpainting with Depth-Guided Cross-View Consistency (3DGIC), achieving multi-view consistent 3D inpainting results with high fidelity.
\item By inferring Depth-Guided Inpainting Masks, the region to be inpainted is properly obtained by considering depth information across different views, allowing us to guide the inpainting process for 3DGS.
\item Based on the 2D inpaintings from a chosen reference view, our Inpainting-guided 3DGS Refinement optimizes new Gaussians of the object-removed scene by ensuring cross-view consistent inpainting results.
\end{itemize}


\section{Related Works}
\label{sec:related}

\begin{figure*}[tb]
	\centering
	\includegraphics[width=1.0\textwidth]{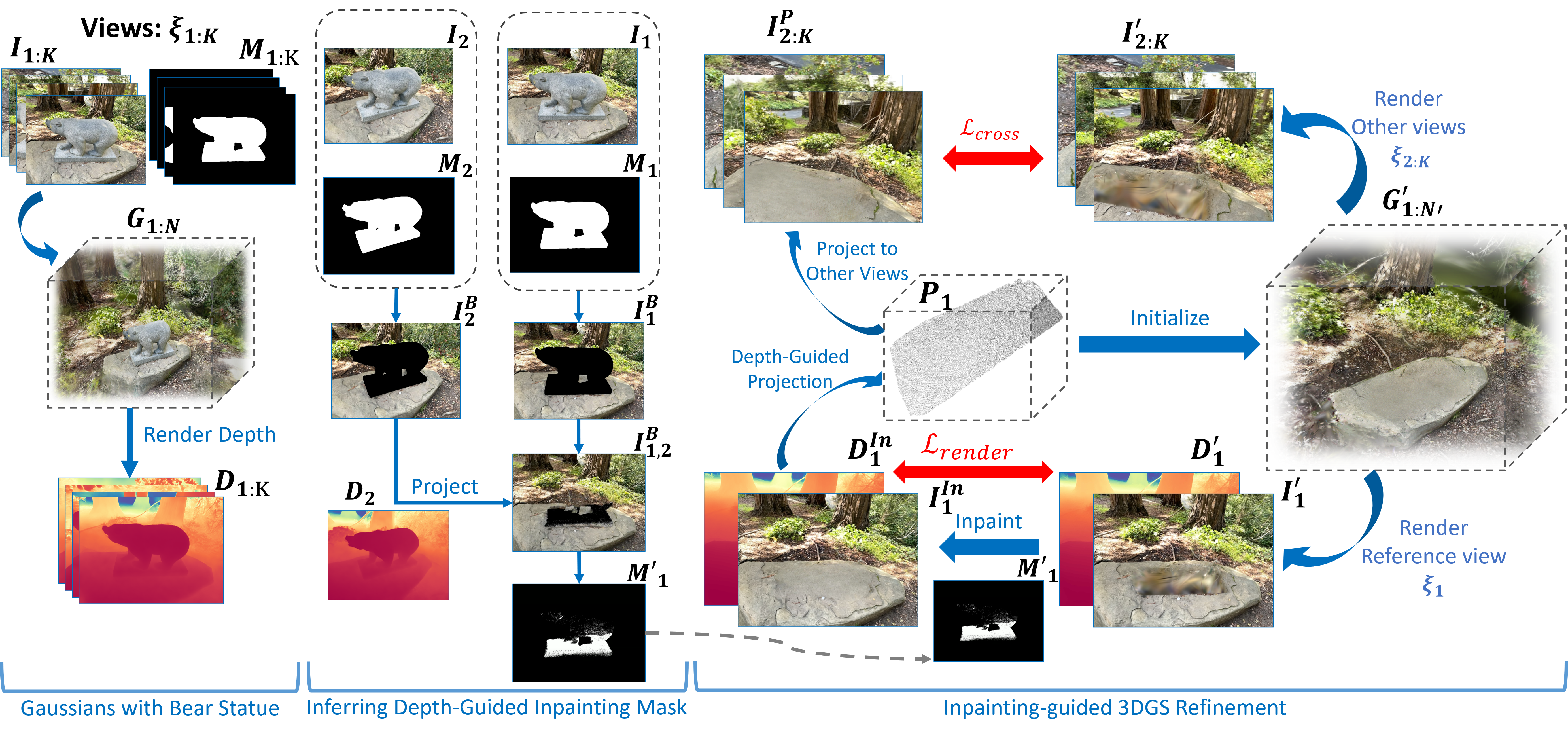}
    \caption{\textbf{Overview of 3D Gaussian Inpainting with Depth-Guided Cross-View Consistency.} Given a 3D Gaussian Splatting model $G_{1:N}$ pretrained on multi-view images $I_{1:K}$ at camera poses $\xi_{1:K}$, our goal is to perform 3D inpainting based on the object masks $M_{1:K}$ (e.g., provided by SAM). With the rendered depth maps $D_{1:K}$, the stage of Inferring Depth-Guide Inpainting Mask is able to refine the inpainting masks to preserve visible backgrounds across camera views. The stage of Inpainting-guided 3DGS Refinement then utilizes such masks to jointly update the new Gaussians $G'_{1:N'}$ for both novel-view rendering and inpainting purposes. 
    }
    \vspace{-1mm}
	\label{fig:main}
\end{figure*}

\subsection{3D Representations for Novel View Synthesis}
\label{subsec:NVS}
Novel view synthesis is a widely studied topic in 3D computer vision. Neural Radiance Field (NeRF)~\cite{mildenhall2021nerforiginal}, a pioneer in this field, effectively models scenes using multi-view images. However, as noted in~\cite{deng2022depth}, the original NeRF requires extensive training time—from hours to days—and relies on numerous images. To address these issues, many subsequent works~\cite{muller2022instant, sun2022direct, yu2021plenoctrees, fridovich2022plenoxels} have emerged. Methods like Instant NGP~\cite{muller2022instant} and DVGO~\cite{sun2022direct} reduce training time to minutes by balancing speed and memory through hash encoding and voxel encoding. Recently, the introduction of 3D Gaussian Splatting (3DGS)~\cite{kerbl202333dgs} brings a fundamental revolution to this area. Different from NeRF and its variants, which model a 3D scene as an implicit representation, 3DGS models a 3D scene as a composition of numerous 3D Gaussians, with each Gaussian parameterized by its three-dimensional centroid, standard deviations, orientations, opacity, and color features. By modeling a 3D scene as such an explicit representation, one is able to render the 2D images of the modeled scene via rasterization with an incredible 100 fps, whereas the fastest NeRF-based approach (\cite{fridovich2022plenoxels, muller2022instant}) only achieves around 10 fps. As a result, we chose 3DGS as our backbone representation over NeRF in this paper due to its fast rendering property, making our approach more applicable in the real world.

\subsection{3D Scene Inpainting}
\label{subsec:3Dinpaint}
In the context of 3D scene inpainting, SPIn-NeRF~\cite{mirzaei2023spin} emerges as one of the earliest approaches addressing the challenges of multi-view consistency. It uses pre-trained segmentation networks to generate plausible inpainting masks for multi-view images, requiring sparse user annotations to indicate the unwanted object. These annotations are propagated across views, and a modified Neural Radiance Field (NeRF) model is used to inpaint the masked regions. Although effective, this approach is heavily dependent on human intervention and lacks the ability to automate the mask generation process, thus limiting its scalability.

To reduce the need for manual annotations, recent works~\cite{yin2023ornerf, ye2023gaussiangrouping} have introduced the use of the Segment Anything Model (SAM)~\cite{kirillov2023sam} in combination with NeRF or 3DGS. Specifically, OR-NeRF employs Grounded-SAM~\cite{ren2024grounded} to locate a single-view 2D inpainting mask for the object to be removed. It then projects 3D points of the object's surface into other views, which are used as prompts for SAM to generate masks for the remaining views. 
Similarly, Gaussian Grouping~\cite{ye2023gaussiangrouping} enhances 3DGS by incorporating semantic feature learning, allowing the model to jointly render RGB images and segmentation maps, where the segmentation supervision is derived from SAM. 
While these methods significantly reduce the burden of manual mask creation, they inpaint 2D images of different views separately and optimize the inpainted NeRF by treating all the 2D inpaintings equally. As a result, the above approaches still face difficulties in producing consistent multi-view results, as mentioned in~\cite{lin2024maldnerf, chen2024mvip, wang2024gscream}

To alleviate this problem, more advanced approaches~\cite{chen2024mvip, lin2024maldnerf, wang2024gscream} focus on improving cross-view consistency. For instance, MALD-NeRF fine-tunes a scene-specific Low-Rank Adaptation (LoRA)~\cite{hu2021lora} module for a pre-trained diffusion model to inpaint images of each scene. By introducing a LoRA module for each scene, the diffusion model can inpaint more consistent content across different views. GScream~\cite{wang2024gscream}, on the other hand, applies diffusion-based 2D inpainting on a chosen reference view. By predicting the depth map of the inpainted reference view, GScream incorporates cross-view feature consistency between any other view and the reference view, optimizing geometric alignment across views. These methods represent a significant step forward in achieving automatic, consistent 3D inpainting, addressing the practical limitations of earlier approaches. Nonetheless, the aforementioned methods rely on per-view 2D inpainting masks for 2D inpainting models as input, while some areas in those masks are visible from other views, as noted in~\cite{ye2023gaussiangrouping}. Consequently, the inpainted content for these visible areas may not align with the original scene (as illustrated in the red branch in Figure 1). This inconsistency might be propagated to the inpainted 3D scene, hindering the reliability of their results.

\section{Method}
\label{sec:method}

\subsection{Problem Definition and Model Overview}
We begin with the notations and settings of our proposed framework. Given a 3D Gaussian Splatting (3DGS) model~\cite{ye2023gaussiangrouping} $G_{1:N} = \{G_1, G_2, ..., G_N \}$ ($N$ denotes the number of Gaussians) pretained for $K$ multi-view images $I_{1:K} = \{I_1, I_2, ..., I_K\}$ with their camera poses $\xi_{1:K} = \{\xi_1, \xi_2, ..., \xi_K\}$, our goal is to remove the Gaussians corresponding to a particular object (e.g., the bear statue) described by 2D object masks $M_{1:K} = \{M_1, M_2, ..., M_K\}$. More precisely, we aim to update the above 3DGS so that the optimized Gaussians $G'_{1:N'}$ (with $N'$ remaining Gaussians) allow novel view rendering without the object of interest presented. Take Figure~\ref{fig:main} as an example, the bear statue is to be removed from the scene of interest, and its segmentation masks $M_{1:K}$ from $I_{1:K}$ can be produced by models like SAM~\cite{kirillov2023sam} (see supplementary materials for details). 


To address the above 3D Gaussian Inpainting with Depth-Guided Cross-View Consistency (3DGIC). Our 3DGIC comprises two learning stages: \textbf{Depth-Guided Inpainting Mask} and \textbf{Inpainting-guided 3DGS Refinement}. The former refines the object mask guided by both semantics and depth maps observed across $I_{1:K}$, while the latter performs inpainting with cross-view consistency for updating the Gaussians $G'_{1:N'}$.

\begin{figure}[t]
	\centering
	\includegraphics[width=1.0\linewidth]{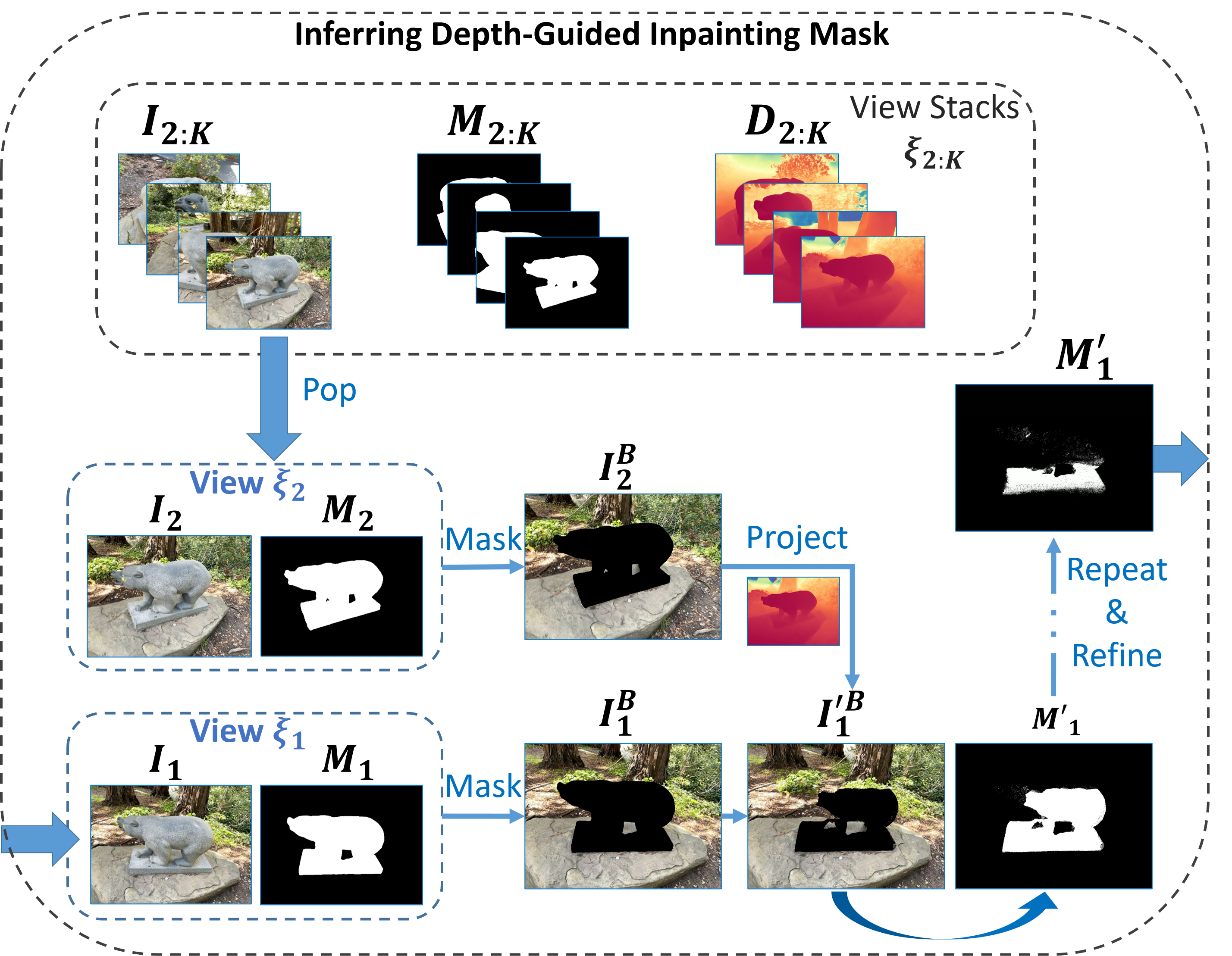}
    \caption{\textbf{Inferring Depth-Guided Inpainting Mask.}  
    Taking $\{I_1, M_1\}$ at view $\xi_1$ as an example reference view, the original background region $I^B_1$ can be first produced. We then project the background region $I^B_2$ from $\xi_2$ to $\xi_1$, updating ${I'}^B_1$ and the associated inpainting mask $M'_1$. By repeating this process across camera views, the final inpainting mask $M'_1$ contains only the regions that are \textit{not} visible at any training camera views.}
    \vspace{-1mm}
	\label{fig:mask_reduct}
\end{figure}


\subsection{Inferring Depth-Guided Inpainting Masks}
\label{subsub:mask reduce}
Given multi-view images $I_{1:K}$ of a scene with binary masks $M_{1:K}$ depicting the object to be removed, we aim to infer a proper mask $M'$ for inpainting images at each view under the guidance of depth images $D_{1:K}$ rendered from $G_{1:N}$. As a result, the masked image $I'^B_{1:K}$
only contains background pixels that are visible at other camera views. The $i$-th masked image $I'^B_i$ is defined as:
\begin{equation}
    I'^B_i = I_i \cdot (\mathds{1}-M_i),
\end{equation}
where $\mathds{1}$ denotes a tensor with the same size as $M$ and all the elements are one.


Take $\{I_1, M_1\}$ in Figure~\ref{fig:mask_reduct} as an example, our process of inferring Depth-Guided Inpainting Masks takes the original image $I_1$ from $\xi_1$ and masks out the areas in $M_1$ as the original visible backgrounds $I^B_1 = I_1\cdot(\mathds{1}-M_1)$ from view $\xi_1$. To explore all the visible background pixels from other views $\xi_{2:K}$, we take $I_{2:K}$ with their masks $M_{2:K}$ and rendered depth $D_{2:K}$ at $\xi_{2:K}$, and we project the above background pixels from each view to $\xi_1$. Taking view $\xi_2$ as an example, the visible backgrounds $I^B_2$ in $I_2$ ($I^B_2 = I_2 \cdot (\mathds{1}-M_2)$) are projected into the 3D space via $D_2$ and $\xi_2$ and then back-projected to $\xi_1$. 
Among all the back-projected pixels, we consider the pixel coordinates that lie inside $M_1$ as visible backgrounds from $I_2$, denoted as $I^B_{1,2}$. The operation for obtain $I^B_{1,2}$ is calculated as:
\begin{equation} \label{eq:proj}
    I^B_{1,2} =  Proj^{2D}(Proj^{3D}(I^B_2, D_2, \xi_2), \xi_1)\cdot M_1,
\end{equation}
where $Proj^{3D}(\cdot, \cdot, \cdot)$ denotes the 3D projection function that projects 2D colored pixels in $I^B_2$ into 3D point clouds with its depth map $D_2$ and camera pose $\xi_2$, while $Proj^{2D}(\cdot, \cdot)$ represents the 2D projection function that projects the 3D colored point cloud back to $\xi_1$ as colored pixels. With the above operation, the corresponding pixel coordinates of $I^B_{1,2}$ are directly excluded from $M_1$, and thus the inpainting mask is refined as $M'_{1}$, and the masked image ${I'}^{B}_1 = I^B_1 + I^B_{1,2}$ at $\xi_1$ is further obtained. Similarly, we repeat this process through all the views $\xi_{2:K}$ to infer the final Depth-Guided Inpainting Mask $M'_1$ and the masked image ${I'}^B_1$ at $\xi_1$. Also, we produce the depth-guided inpainting masks $M'_{1:K}$ for all the views $\xi_{1:K}$. Please refer to our supplementary material for more details about this inferring process. 

It is worth noting that the above process is deterministic. It not only reduces the uncertainty of image regions to inpaint at each view, but it also makes the updating of the 3DGS model for rendering the unpainted scene more effective, as discussed in the following subsection.

\subsection{Inpainting-guided 3DGS Refinement}
\label{subsub:optimize}
The aim of this stage is to optimize $G'_{1:N'}$ with masked $I'_{1:K}$ obtained at $\xi_{1:K}$ with cross-view consistency so that rendering of the corresponding high-fidelity scene can be produced, realizing the task of 3D inpainting. As shown in Figure~\ref{fig:main}, the 3DGS for this inpainting scene can be first updated by removing the Gaussians with the semantic labels corresponding to the masked region (e.g., ``\textit{bear}'' in the original Gaussian $G_{1:N}$), and replaced by the same amount of randomly initialized Gaussians in the masked region (e.g., with bear removed; see~\cite{ye2023gaussiangrouping} and our supplementary materials). 

Take $\xi_1$ as the reference view for an example, the rendered image $I'_1$ and depth map $D'_1$ of $G'_{1:N'}$ at $\xi_1$ are inpainted by a 2D inpainter~\cite{lama, ldm} (using $M'_1$ as inpainting mask) as:

\begin{equation}
\label{eq:lrender}
\begin{aligned}
    I^{In}_1 & =  & Inpaint_{2D}(I'_1, M'_1) \\
    D^{In}_1 & = & Inpaint_{2D}(D'_1, M'_1),
\end{aligned}
\end{equation}
where $Inpaint_{2D}(\cdot, \cdot)$ denotes the 2D inpainting process, and $I^{In}_1$ and $D^{In}_1$ represents the 2D-inpainted results of $I'_1$ and $D'_1$, respectively.
To ensure $I'_1$ looks identical to $I^{In}_1$, the \textit{rendering loss} at $\xi_1$ is defined as:
\begin{equation}
\label{eq:lrender}
    \mathcal{L}_{rendering} = \mathcal{L}_{rgb} + \mathcal{L}_{depth}.
\end{equation}
Note that the image recovery loss $\mathcal{L}_{rgb}$ is calculated as:
\begin{equation}
\label{eq:optimageloss}
    \mathcal{L}_{rgb} = \| I'_1 - {I^{In}_1}\|_1 + \mathcal{L}_{SSIM}(I'_1, I^{In}_1),
\end{equation}
where the $\mathcal{L}_{SSIM}$ denotes the structure similarity loss~\cite{kerbl202333dgs}. And, the depth loss $\mathcal{L}_{depth}$ is defined as:
\begin{equation}
\label{eq:depthloss}
    \mathcal{L}_{depth} = \| D'_1 - {D^{In}_1}\|_1.
\end{equation}

To further ensure the masked regions in $I'_{2:K}$ (with respect to $M'_{2:K}$) are cross-view consistent with the 2D-inpainted region in $I^{In}_1$, we project the inpainted region of $I^{In}_1$ into the 3D space as a set of colored point clouds $P_1$, followed by re-projecting back to $\xi_{2:K}$ as supervision. Thus, $P_1$ is calculated as:
\begin{equation}
    P_1 =  Proj^{3D}(I^{In}_1 \cdot M'_1, D^{In}_1, \xi_1)),
\end{equation} 
where $Proj^{3D}(\cdot, \cdot, \cdot)$ is the same projection function in Eqn.~\ref{eq:proj}. For each view $\xi_k$  of $\xi_{2:K}$, the back-projected image $I^P_k$for supervision is denoted as:
\begin{equation}
    I^{P}_k =  I'_k \cdot (1-M'_k) + Proj^{2D}(P_1, \xi_k)\cdot M'_k,
\end{equation}
where $Proj^{2D}(\cdot, \cdot)$ is also the same 2D projection function in Eqn.~\ref{eq:proj}. To this end, the cross-view consistent loss $\mathcal{L}_{cross}$ is defined as:
\begin{equation} \label{Lcross}
    \mathcal{L}_{cross} = \sum_{k\in {2...K}} \mathcal{L}_{LPIPS}(I'_k, I^{P}_k),
\end{equation}
where $\mathcal{L}_{LPIPS}$ denotes the LPIPS~\cite{lpips} loss that calculates the perceptual similarity between $I'_k$ and $I^{P}_k$.

Finally, the overall loss for 3D inpainting is calculated by $\mathcal{L}_{inpaint} = \mathcal{L}_{render} + \mathcal{L}_{cross}$. We note that by conducting $\mathcal{L}_{inpaint}$, $G'_{1:N'}$ is guaranteed to inpaint the object-removed 3D scene with cross-view consistency by taking $\{I^{In}_1, D^{In}_1 \}$ as guidance. 


\subsection{Training and Inference}
\label{subsec:train and inference}
\subsubsection{Training}
During the training (optimization) process, we calculate the refined mask $M'$ described in Sect.~\ref{subsub:mask reduce} for all $K$ views and choose the view with the largest refined mask as the reference view. This is because the 2D inpainted result from this view covers the most 3D space compared to other views, allowing us to provide a more informative cross-view optimization. By choosing the reference view, $\mathcal{L}_{inpaint}$ is applied to optimize $G'_{1:N'}$. To this end, $G'_{1:N'}$ is properly supervised to ensure the 3D scene is reasonably inpainted and consistent across different views. 

\subsubsection{Inference}
Once we finish the optimization of the inpainted scene with our 3DGIC, the optimized Gaussians  $G'_{1:N'}$ are able to render a novel view synthesis of the scene by using arbitrary camera poses.


\section{Experiments}
\label{sec:exp}
\begin{table*}[ht]
\centering
 \caption{\textbf{Quantitative evaluation on the SPIn-NeRF dataset in terms of FID and LPIPS.} Note that m-FID and m-LPIPS represent that the FID and LPIPS scores are only calculated within the ground truth inpainting masks.}
 \label{tab:main_table}
\resizebox{0.7\textwidth}{!}{%
\begin{tabular}{l|c|c|c|c|c|c}
                         & Representation &2D inpainter & FID$\downarrow$ & m-FID$\downarrow$ & LPIPS$\downarrow$ & m-LPIPS$\downarrow$ \\ \hline
SPIn-NeRF~\cite{mirzaei2023spin} &NeRF               & LAMA~\cite{lama}         & 49.6& 153.4  &0.31   & 0.053  \\
MVIP-NeRF~\cite{chen2024mvip} &NeRF               & LDM~\cite{ldm}         & 50.5& 173.4  &0.31   & 0.050  \\
Gaussian Grouping~\cite{ye2023gaussiangrouping} & Gaussian Splatting      & LAMA~\cite{lama}         & 44.7& 132.5  &0.30   & 0.037  \\
MALD-NeRF~\cite{lin2024maldnerf} & NeRF                & LDM~\cite{ldm}          & 44.9& 113.5  &0.26   &0.031   \\
GScream~\cite{wang2024gscream}   & Gaussian Splatting              & LDM~\cite{ldm}          & 38.6& 101.6  &0.28   &0.033   \\ \hline
3DGIC (Ours)             & Gaussian Splatting        & LAMA~\cite{lama}         & 41.7& 102.4  &0.28   &0.032   \\
3DGIC (Ours)              & Gaussian Splatting       & LDM~\cite{ldm}          & \textbf{36.4}& \textbf{96.3}   &\textbf{0.26}   &\textbf{0.028}    \\                    
\end{tabular}}
\end{table*}
\begin{figure*}[tb]
	\centering
	\includegraphics[width=1.0\textwidth]{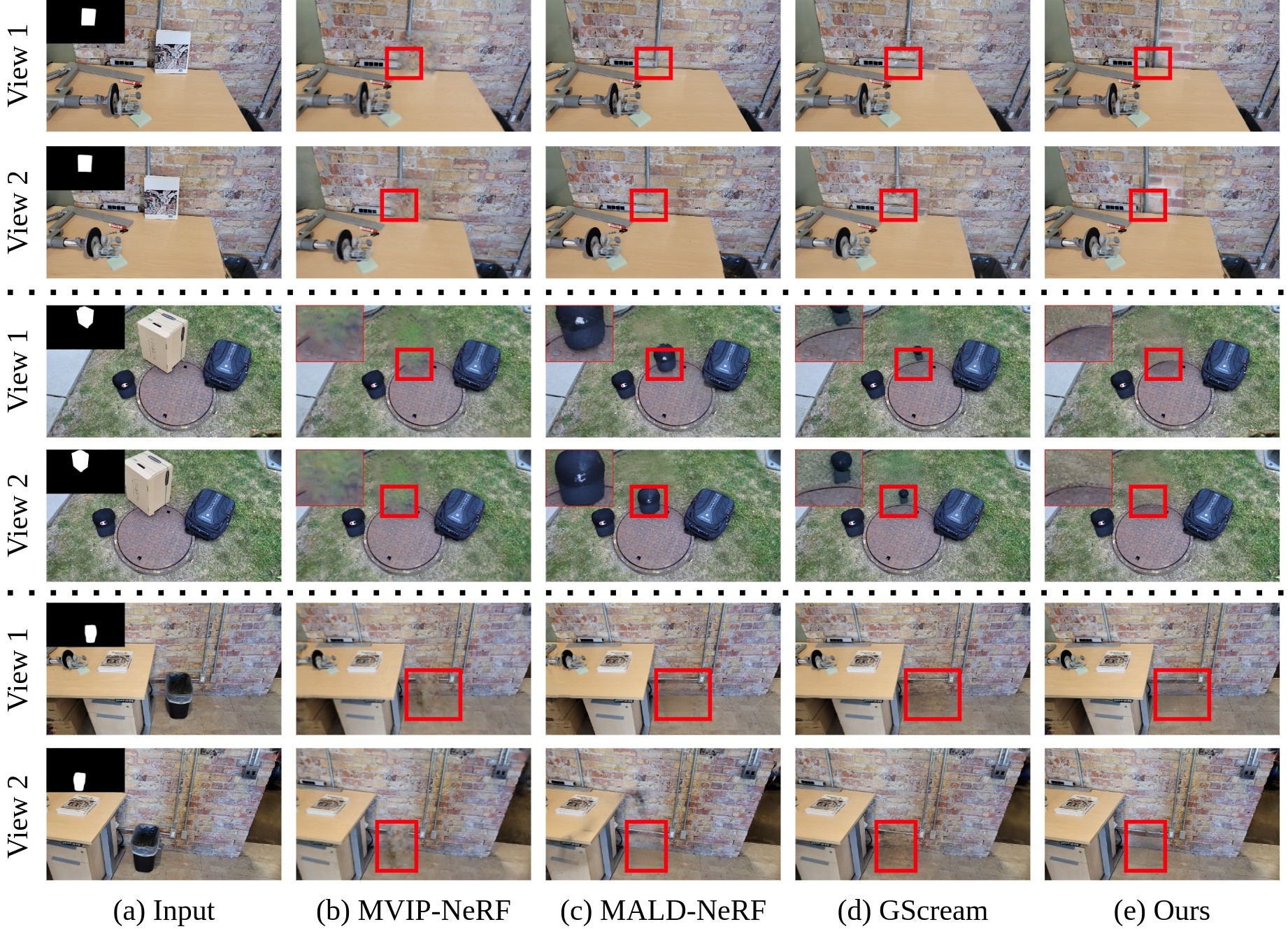}
    \caption{\textbf{Qualitative results on the SPIn-NeRF~\cite{mirzaei2023spin} dataset.} Two different views of the same scene are shown for each inpainting example. We compare rendering results against MVIP-NeRF~\cite{chen2024mvip}, MALD-NeRF~\cite{lin2024maldnerf}, and GScream~\cite{wang2024gscream}. We can see from the regions highlighted by the red boxes that our 3DGIC performs better in terms of multi-view consistency and rendering fidelity
    }
    \vspace{-1mm}
	\label{fig:quali_spin}
\end{figure*}

\begin{figure*}[tb]
	\centering
	\includegraphics[width=1.0\textwidth]{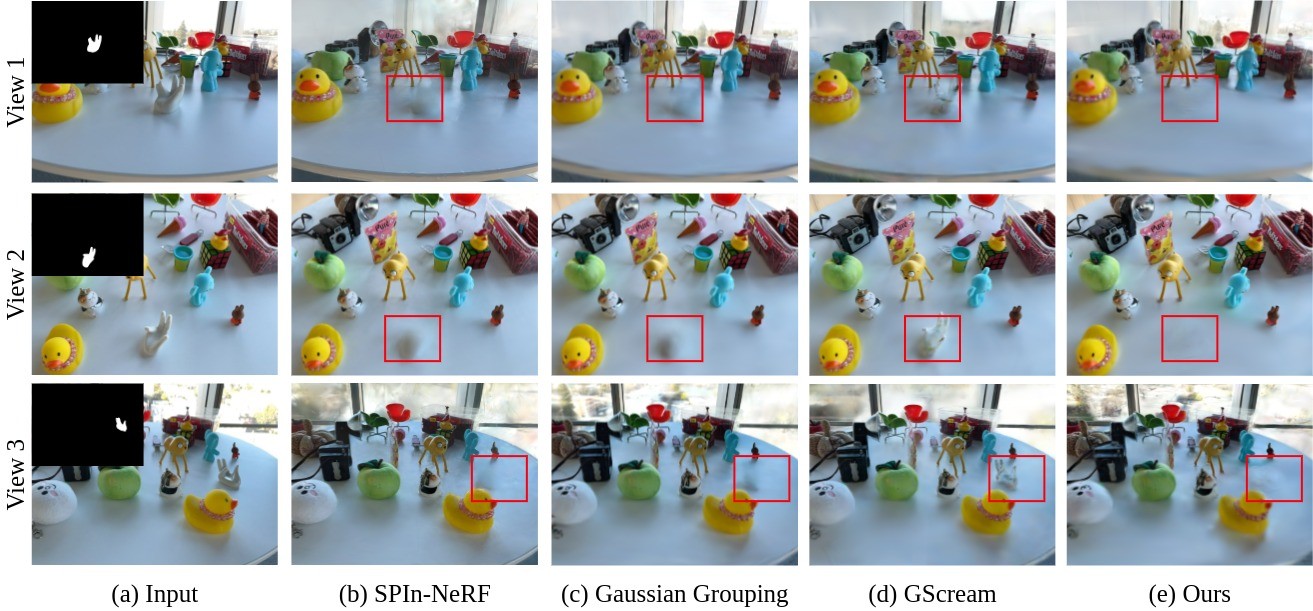}
    \caption{\textbf{Qualitative results on the \textit{Figurines} scene from the LeRF~\cite{kerr2023lerf} dataset.} We compare the rendering results with SPIn-NeRF~\cite{mirzaei2023spin}, Gaussian Grouping~\cite{ye2023gaussiangrouping}, and GScream~\cite{wang2024gscream}. The three rows show different views of the scene, whereas the first column shows the input images with the object masks of the unwanted object. The regions highlighted by the red boxes show that our 3DGIC inpaints a smoother table surface without artifacts.   
    }
    \vspace{-1mm}
	\label{fig:quali_figurines}
\end{figure*}

\begin{figure*}[tb]
	\centering
	\includegraphics[width=1.0\textwidth]{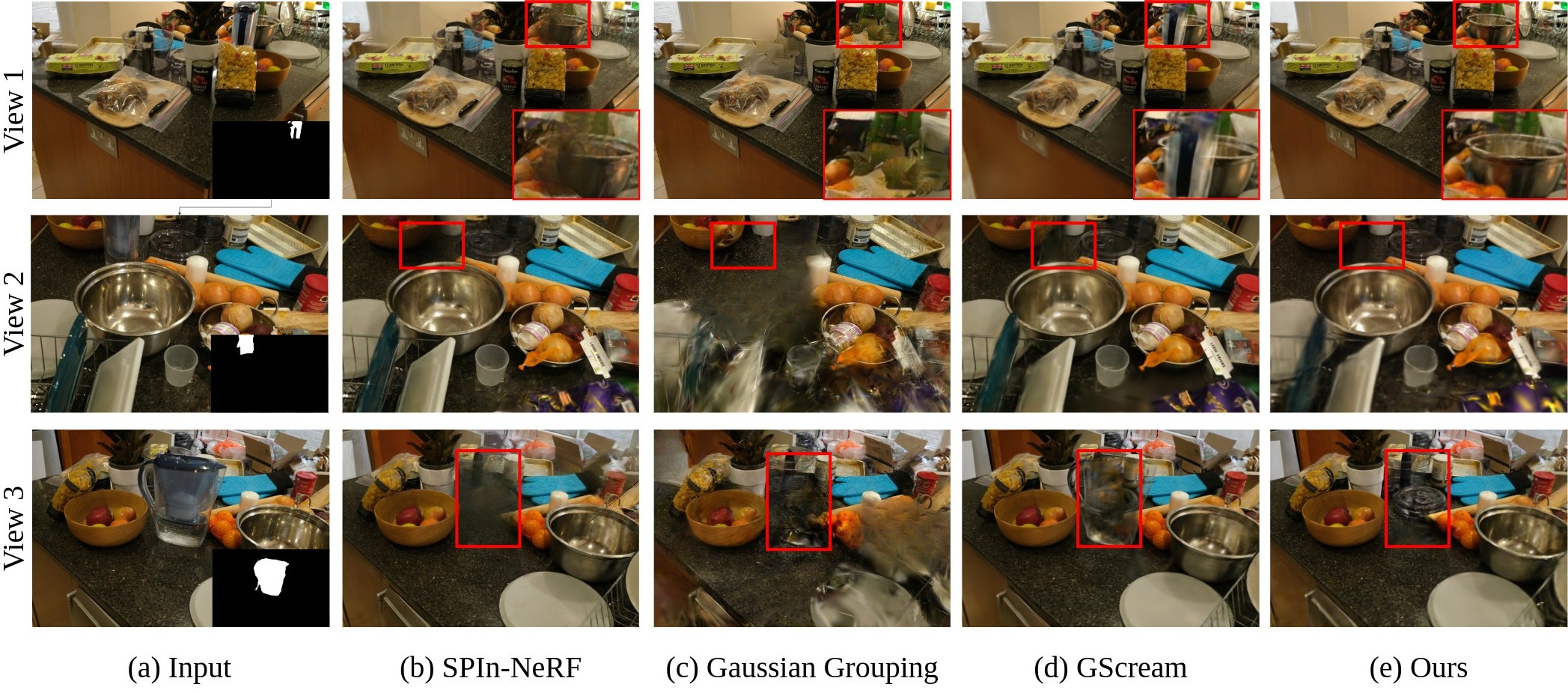}
    \caption{\textbf{Qualitative results on the \textit{Counter} scene from the MipNeRF360~\cite{barron2022mipnerf360} dataset.} We compare the rendering results with SPIn-NeRF~\cite{mirzaei2023spin}, Gaussian Grouping~\cite{ye2023gaussiangrouping}, and GScream~\cite{wang2024gscream}. The three rows show different views of the scene, where we zoom in a certain region in the first row to highlight the difference between each method. We can see from the regions highlighted by the red boxes that our 3DGIC correctly inpaints the water bottle without manipulating any other objects on the table (e.g., the plastic cover).
    }
    \vspace{-1mm}
	\label{fig:quali_counter}
\end{figure*}

\begin{figure*}[tb]
	\centering
	\includegraphics[width=1.0\textwidth]{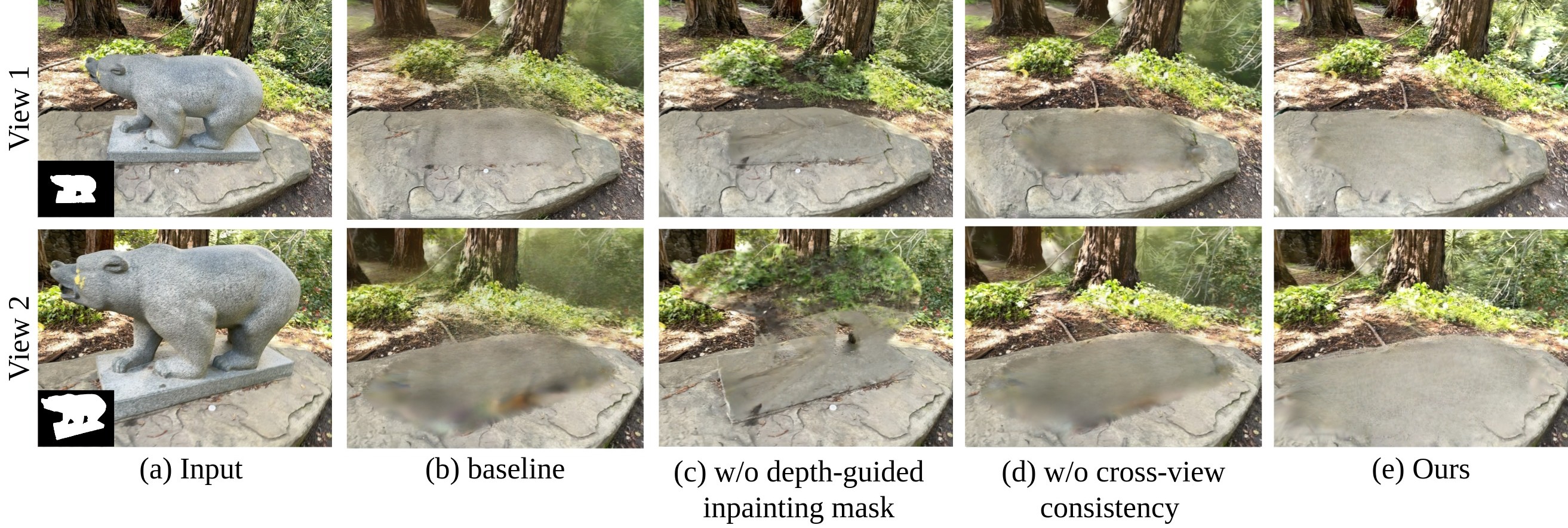}
    \caption{\textbf{Ablation studies on the \textit{Bear} scene from the InNeRF360~\cite{wang2024innerf360} dataset.} We verify the effectiveness of our Inferring Depth-Guided Inpainting Mask and Inpainting-guided 3DGS Refinement.
    }
    \vspace{-1mm}
	\label{fig:ablation}
\end{figure*}

\subsection{Datasets}
To evaluate the effectiveness of our method, we conduct experiments on the most used real-world benchmark dataset: the SPIn-NeRF~\cite{mirzaei2023spin} dataset. This dataset contains \textit{ten} real-world scenes, including indoor and outdoor scenes. Each scene is composed of 60 frames of training images and 40 frames of testing images where a certain object in the scene is removed, with camera poses of all 100 images available. The binary mask of the object to be removed is also provided in each frame for evaluation. Following the setting of \cite{mirzaei2023spin, lin2024maldnerf, wang2024gscream, ye2023gaussiangrouping, chen2024mvip}, we resize each image as $1008 \times 567$ in resolution for all our experiments and show the comparisons quantitatively and qualitatively. 

Since the camera poses in all the scenes provided in the SPIn-NeRF dataset only cover a small range (i.e., all the image frames are captured near the front view of the scene), we additionally include qualitative comparisons with several scenes covering 360$^{\circ}$ of camera poses to show the effectiveness of our design, specifically for our Depth-Guided Inpainting Mask. Following Gaussian Grouping~\cite{ye2023gaussiangrouping}, we take the ``\textit{bear}'' scene provided in InNeRF360~\cite{wang2024innerf360}, the ``\textit{counter}'' scene in Mip-NeRF360~\cite{barron2022mipnerf360}, and the ``\textit{figureines}'' scene in LeRF~\cite{kerr2023lerf} for the additional qualitative evaluations. Since these scenes are not originally for the 3D inpainting task, we manually select an object in each scene as the object to be removed and select the corresponding ID in the segmentation map obtained from SAM~\cite{kirillov2023sam} as the object mask in each view. Please refer to our supplementary material for a detailed description of these scenes.

\subsection{Quantitative Evaluations}
\label{sec:quant}
Table~\ref{tab:main_table} shows the comparisons between our 3DGIC (with LAMA~\cite{lama} or LDM~\cite{ldm} as 2D inpainter) and several state-of-the-art approaches such as SPIn-NeRF~\cite{mirzaei2023spin}, MVIP-NeRF~\cite{chen2024mvip}, Gaussian Grouping~\cite{ye2023gaussiangrouping}, MALD-NeRF~\cite{lin2024maldnerf}, and GScream~\cite{wang2024gscream} using the SPIn-NeRF dataset. Following SPIn-NeRF and MALD-NeRF, we conduct FID~\cite{heusel2017fid}, masked FID (m-FID), LPIPS~\cite{lpips}, and masked LPIPS (m-LPIPS) as our evaluation matrices, where m-FID and m-LPIPS calculate the FID and LPIPS scores only inside the ground truth inpainting masks. We note that the official implementation of MALD-NeRF is currently unavailable; we directly use the output results provided on their official project page for evaluation. As for other state-of-the-arts, we reproduce results from their official implementations and the released configurations.

From Table~\ref{tab:main_table}, we can see that the LDM version of our 3DGIC achieves the best score on all four evaluation matrices. As for our 3DGIC using a non-diffusion-based model of LAMA as the 2D inpainter, the results still outperform MVIP-NeRF and MALD-NeRF, where both use LDM as the inpainter. The above results show that while using a better 2D inpainter achieves better results, the improvements in our 3DGIC do not come solely from a better 2D inpainter. This suggests that our model is not bundled by 2D inpainters and achieves 3D inpainting with improved fidelity.

\subsection{Qualitative Results}
In Figure~\ref{fig:quali_spin}, we qualitatively compare our 3DGIC with MVIP-NeRF~\cite{chen2024mvip}, MALD-NeRF~\cite{lin2024maldnerf}, and Gscream~\cite{wang2024gscream} using the testing set of SPIn-NeRF dataset. In this figure, each of the two rows shows the results of the same scene with different viewpoints, while the first column shows the images containing the object to be removed along with the object masks at the upper-left corner. Specifically, from the first two rows, we observe that while GScream and MALD-NeRF both show high-fidelity images, some of the visible details from the input image (e.g., the electrical socket on the table) are not preserved properly. For the third and fourth rows, where we zoom in on certain areas inside the red boxes, although it is reasonable for MALD-NeRF to generate a hat in the inpainted region, the logo on the hat is not consistent across different views. As for MVIP-NeRF, blurry images are generated in all cases. Oppositely, our 3DGIC generates high-fidelity images with multi-view consistency and preservation of the visible backgrounds. 

In Figure~\ref{fig:quali_figurines} and Figure~\ref{fig:quali_counter}, we further show the qualitative comparisons with SPIn-NeRF, Gaussian Grouping, and GScream using the \textit{Figurines} dataset from LeRF~\cite{kerr2023lerf} and the \textit{Counter} dataset from MipNeRF360~\cite{barron2022mipnerf360}, where each shows results from three different viewpoints. For Figure~\ref{fig:quali_figurines}, we can see that both SPIn-NeRF and Gaussian Grouping leave obvious black holes and shadows in the inpainting region, while GScream does not clearly remove the object of interest. In contrast, our 3DGIC successfully removes the unwanted object and produces smooth and multi-view consistent results without leaving heavy shadows. For Figure~\ref{fig:quali_counter}, where certain areas are cropped by red boxes and zoomed in in the first row, GSream does not fully remove the object of interest either. SPIn-NeRF not only removes the object of interest but also inpaints other objects in the background. As for Gaussian Grouping, which uses GroundedSAM~\cite{ren2024grounded} to detect inpainting mask with the text prompt ``\textit{blurry hole}'' as input, the GroundedSAM model locates other regions rather than focusing on the object removed region, producing blurry and inconsistent inpainting results across different views. In the contrary, our 3DGIC locates the regions to be inpaint properly and hence produces high-fidelity results while preserving all the other background objects.

\subsection{Ablation Study}
\label{sec:ablation}
To further analyze the effectiveness of our designed modules (i.e., Inferring Depth-Guided Masks and Inpainting-guided 3DGS Refinement), we conduct ablation studies on the ``\textit{bear}'' scene from InNeRF360~\cite{wang2024innerf360}, as shown in Figure~\ref{fig:ablation}. Column (a) shows the input images with the bear statue and their corresponding object mask. The baseline model (b) uses the original object masks as the inpainting masks and directly applies all the inpainted 2D images as input to fine-tune a 3DGS model. The results of model (b) show blurry contents all over the rendered image, while the inpainted results are not consistent across different views.  For model (c), the original object masks are applied as the 2D inpainting model, with our Inpainting-guided 3DGS Refinement. Although the rendered images of model (c) show better fidelity, using the original object masks as inpainting masks results in modifications to the visible backgrounds. For model (d), our inferred depth-guided masks $M'_{1:K}$ are applied as the 2D inpainting masks, but all the 2D inpainting results are directly used as inputs to fine-tune the 3DGS model. As a result, although the backgrounds are preserved, the inpainted region is blurry and not consistent across the views. As for our full model in the last column (e), the depth-guided masks are used, and the 3D Inpainting with Cross-View Consistency is applied, achieving the best results. This verifies the success of our proposed modules and strategies for 3D inpainting.

\section{Conclusions}
In this paper, we propose the 3D Gaussian Inpainting with Depth-Guided Cross-View Consistency (3DGIC) for inpainting real-world 3D scenes represented by 3D Gaussian Splatting (3DGS) models. With the conduction of our Inferring Depth-Guided Inpainting Masks, we are allowed to obtain precise inpainting masks by considering rendered depth maps and visible background information from other views. With these depth-guided inpainting masks properly obtained, our Inpainting-guided 3DGS Refinement optimizes a newly initialized 3DGS model and performs 3D inpainting simultaneously. In our experiments, we quantitatively and qualitatively show that our 3DGIC is able to handle scenes with various ranges of camera views and perform favorably against existing 3D inpainting approaches.

\renewcommand{\thefigure}{A\arabic{figure}}
\renewcommand{\thetable}{A\arabic{table}}

\clearpage
\setcounter{page}{1}
\maketitlesupplementary
\appendix

\section{Additional Details of 3DGIC}
\label{sec:details}

\subsection{Details of Backbone 3D Gaussian Splatting Model}
\label{subsec:backbone}
Given the multi-view images $I_{1:K}$ with corresponding camera poses $\xi_{1:K}$ of a 3D scene, the vanilla 3DGS~\cite{kerbl202333dgs} model parameterize each Gaussian $G_i$ in $G_{1:N}$ with its 3-dimensional centroid $\mathbf{p}_i \in \mathbb{R}^{3}$, a 3-dimensional standard deviation $\mathbf{s}_i \in \mathbb{R}^{3}$, a 4-dimensional rotational quaternion $\mathbf{q}_i \in \mathbb{R}^{4}$, an opacity ${\alpha}_i \in [0,1]$, and color coefficients $\mathbf{c}_i$ for spherical harmonics in degree of 3. Hence, $G_i$ is represented with a set of the above parameters (i.e., $G_i = \{\mathbf{p}_i, \mathbf{s}_i, \mathbf{q}_i, {\alpha}_i, \mathbf{c}_i\}$). However, to make sure the 3DGS models in this paper are capable of removing Gaussians corresponding to any indicated object (e.g., ``bear'' in Figure \textcolor{cvprblue}{2}) as described in Sect.~\textcolor{cvprblue}{3.3}, we incorporate the use of a semantic-aware 3DGS (i.e., Gaussian Grouping~\cite{ye2023gaussiangrouping}) approach as the main backbone 3DGS model of our method. Also, since the rendered depth maps $D_{1:K}$ are utilized as important guidance in our 3DGIC, we additionally combine the use of Relightable Gaussian~\cite{gao2023relightable}, which produces better depth estimations from 3DGS model as our final backbone for Sect.~\textcolor{cvprblue}{3}. We now briefly discuss both methods.  

\paragraph{Incorporating Semantic Segmentation via Gaussian Grouping.}
To overcome the lack of fine-grained scene understanding in 3DGS, Gaussian Grouping~\cite{ye2023gaussiangrouping} extends 3DGS by incorporating segmentation capabilities. Along with $I_{1:K}$, Gaussian Grouping additionally takes the Segment Anything Model (SAM) to produce 2D semantic segmentation masks $S_{1:K} = \{S_1, S_2, ..., S_K\}$ from multiple views as inputs, and an additional 16-dimensional parameter $\mathbf{e}_i \in \mathbb{R}^{16}$ is introduced to represent a 3D Identity Encoding for each Gaussian $G_i$. Therefore, each Gaussian $G_i$ is extended as $G_i = \{\mathbf{p}_i, \mathbf{s}_i, \mathbf{q}_i, {\alpha}_i, \mathbf{c}_i, \mathbf{e}_i\}$. To make sure $G_{1:K}$ learns to segment each object represented by $S_{1:K}$ in the scene, a 2D identity loss $\mathcal{L}_{id}$ is applied by calculating cross-entropy between $\hat{S}_{1:K}$ and $S_{1:K}$, where $\hat{S}_{1:K} = \{\hat{S}_1, \hat{S}_2, ... , S_K\}$ denotes the rendered segmentation maps from $G_{1:K}$. Additionally, to further ensure that the Gaussians having the same identities are grouped together, a 3D regularization loss $\mathcal{L}_{3D}$ is applied to enforce each $G_i$'s k-nearest 3D spatial neighbors to be close in their feature distance of Identity Encodings. Please refer to the original paper~\cite{ye2023gaussiangrouping} for detailed formulations of segmentation map rendering and $\mathcal{L}_{3D}$. The design of Gaussian Grouping ensures that the segmentation results are coherent across multiple views, enabling the automatic generation of binary masks for any queried object in the scene.

\paragraph{Produce Reliable Depth Estimations with Relightable Gaussians.}
Different from Gaussian Grouping, Relightable Gaussians~\cite{gao2023relightable} extends the capabilities of Gaussian Splatting by incorporating Disney-BRDF~\cite{burley2012brdf} decomposition and ray tracing to achieve realistic point cloud relighting. 
Unlike traditional Gaussian Splatting, which primarily focuses on appearance and geometry modeling, Relightable Gaussians also aim to model the physical interaction of light with different surfaces in the scene.
Specifically, for each Gaussian $G_i$, the original color coefficients $\mathbf{c}_i$ is decomposed into a 3-dimensional base color $\mathbf{b}_i \in [0,1]^3$, a 1-dimensional roughness $r \in [0,1]$, and incident light coefficients $\mathbf{l}_i$ for spherical harmonics in degree of 3. Subsequently, the Physical-Based Rendering (PBR) process and a point-based ray tracing are applied to obtain the colored PBR 2D images $\hat{I}^{PBR}_{1:K}$ and additionally supervised by $I_{1:K}$. Besides the above extensions on PBR for relighting, Relightable Gaussians also introduces a 3-dimensional normal $\mathbf{n}_i$ for $G_i$ and leverages several techniques, including an unsupervised estimation of a depth map $D_i$ from each input view $\xi_i$, to enhance the geometry accuracy and smoothness. By conducting this self-supervised estimation and regularization of normal maps and depth maps, the predicted depth map $D_i$ is more reliable than the vanilla 3DGS. Please refer to the original paper of Relightable Gaussians~\cite{gao2023relightable} for detailed explanations.  

In conclusion, each Gaussian of our 3DGIC is parameterized as $G_i = \{\mathbf{p}_i, \mathbf{s}_i, \mathbf{q}_i, {\alpha}_i, \mathbf{c}_i, \mathbf{e}_i, \mathbf{b}_i, r,  \mathbf{l}_i, \mathbf{n}_i\}$. By combining these methods, we are able to perform reliable depth estimations and effective removal of the Gaussians corresponding to any object in the scene for our 3DGIC.

\subsection{Additional Details of  Inferring Depth-Guided Inpainting Masks}
In Sect.~\textcolor{cvprblue}{2.2} in our main paper, we introduce infer proper inpainting masks $M'_{1:K}$ to determine the region to be inpaint by realizing visible background regions across different views. In our implementation, after updating the inpainting masks $M'_{1:K}$ with the process described in Sect.~\textcolor{cvprblue}{3.2}, we further conduct a refinement for each mask as a post-processing to prevent noisy mask. Taking $M'_1$ as an example, this process updates $M'_1$ as:

\begin{equation}
    M'_1 \leftarrow Open(M'_1), 
\end{equation}
where $Open(\cdot)$ represents a morphological opening process to reduce noises. This refinement process ensures that small noisy pixels are suppressed in our Depth-Guided Inpainting Masks.
\begin{figure*}[tb]
	\centering
	\includegraphics[width=1.0\textwidth]{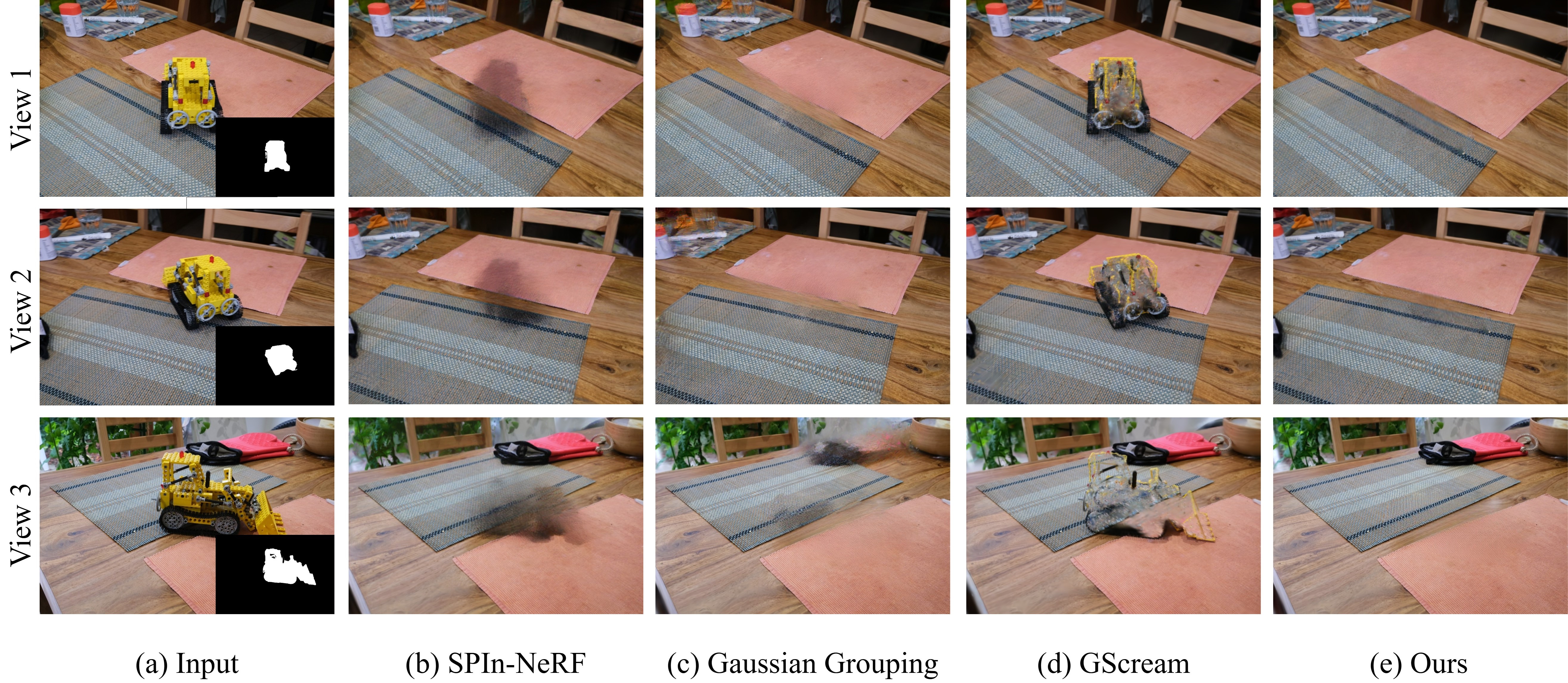}
    \caption{\textbf{Qualitative results on the \textit{Kitchen} scene from the MipNeRF360~\cite{barron2022mipnerf360} dataset.} We compare the rendering results with SPIn-NeRF~\cite{mirzaei2023spin}, Gaussian Grouping~\cite{ye2023gaussiangrouping}, and GScream~\cite{wang2024gscream}. The three rows show different views of the scene. We can see that our 3DGIC inpaint a smooth kitchen table, while other approaches produce blurry results.
    }
    \vspace{-1mm}
	\label{fig:quali_kitchen}
\end{figure*}

\subsection{Additional Details of  Initializing Inpainted Gaussian}
In Sect.~\textcolor{cvprblue}{3.3}, we introduce to remove the Gaussians with semantic labels corresponding to the ``\textit{bear}'' object in $G_{1:N}$ and replace by the same amount of randomly initialized Gaussians in the masked region as the initialization of $G'_{1:N'}$. We now detail this initialization process for $G'_{1:N'}$. 

When first removing the Gaussians corresponding to the ``bear'' object, we directly use the remaining Gaussian to render the image $I'_1$ and depth map $D'_1$. Following the 2D inpainting process described in Sect.~\textcolor{cvprblue}{3.3}, the inpainted image $I^{In}_1$ and depth map $D^{In}_1$ are produced and projected into 3D space as colored point clouds $P_1$. We then use the 3D coordinates of $P_1$ as the initialized 3D position for the newly introduced Gaussians for $G'_{1:N'}$, since $P_1$ represents the ideal surface of the inpainted 3D Gaussian provided by $I^{In}_1$ after removing the bear. Note that if the number of points in $P_1$ does not match the number of newly initialized Gaussians in $G'_{1:N'}$ (also the number of removed Gaussians in $G_{1:N}$), we apply random selection to the coordinates of $P_1$ to match the number of the newly introduced Gaussians. As for the other parameters of the newly introduced Gaussians in $G'_{1:N'}$, we follow Gaussian Grouping~\cite{ye2023gaussiangrouping} to average the parameters of each Gaussian's 5-nearest neighbors (in 3D space) from the remaining Gaussians as initialization. By this process, $G'_{1:N'}$ is properly initialized.

\subsection{Implementation Details}
 In all our experiments, we train one model for each object category, using a single NVIDIA RTX 3090  GPU (24G) for training with the PyTorch~\cite{paszke2019pytorch}  libraries. For each scene, 5000 iterations of optimization are applied to obtain the inpainted 3DGS model. We also use the official implementation of \cite{wang2024gscream, ye2023gaussiangrouping, chen2024mvip, mirzaei2023spin} for comparison. When applying 2D inpainting models to the image and depth map to be inpaint, if we use non-diffusion-based LAMA~\cite{lama} as inpainter, the RGB image and depth map are inpainted separately. However, if LDM~\cite{ldm} is applied as our 2D inpainters, we follow the suggestion in NeRFiller~\cite{weber2024nerfiller} to stack the RGB image and the depth map in the same image for inpainting to ensure the inpainted RGB image and the depth map are consistent in terms of the geometry details. Specifically, we crop a $512 \times 512$ patch for the RGB image and the depth map to be inpainted center at the pixel coordinate of the inpainting mask's center, and paste the cropped RGB patch to a $1024 \times 1024$-resolution black image at the upper right corner with the cropped depth map at the lower left corner as the input image for the LDM. Similarly, we also crop a $512 \times 512$ patch for the inpaint masks and put them to the upper right and lower left corner of another $1024 \times 1024$-resolution black image as the input binary inpainting mask for the LDM. We then use the prompt ``\textit{an RGB image and a depth image of the same scene}'' to inpaint the input image. Finally, the inpainted RGB patch and the depth map patch are pasted back to the original image and depth map, respectively, as the 2D inpainting result. It is worth noting that we apply the 2D inpainting process for every 500 iterations. Following MALD-NeRF~\cite{lin2024maldnerf}, we use the technique of partial DDIM~\cite{song2020ddim}, to start from latter step of the denoising process as optimization iteration grows. Specifically, for a 50-step DDIM process, we start from step 0 of the LDM denoising process for step 0 of our optimization. After 500 iteration steps, the second time of the LDM inpainting starts from step 5 of the DDIM process and so on. When our optimization reaches the last 500 iterations, the 2D inpainting process only denoises using the last five steps of DDIM. This prevents inpainting results that are too different from the current scene and provides more stability for our optimization process.

\subsection{Dataset Details}
For the ``figurines'' scene from LeRF~\cite{kerr2023lerf} dataset, we have 260 training frames and 40 testing frames, each with a resolution of $986 \times 728$. For the ``bear'' dataset from InNeRF360~\cite{wang2024innerf360}, we have 90 training frames and 6 testing frames, each with a resolution of $985\times729$. As for ``counter'' and ``kitchen'' scenes from MipNeRF360~\cite{barron2022mipnerf360}, 240 (230 for training and 10 for testing) and 279 (270 for training and 9 for testing) frames are available in total, respectively. Both scenes are in the resolution of $779\time520$.

\section{Additional Experiments}
We additionally show the results on the ``\textit{kitchen}'' scene from the MipNeRF360~\cite{barron2022mipnerf360} dataset in Figure~\ref{fig:quali_kitchen}. We can see that SPIn-NeRF produces blurry result, while GScream fail to handle camera views with a wide range and not able to remove the excavator clearly. Although Gaussian Grouping also produces plausible results at the excavator-removed regions, it incorrectly detects the glove behind the excavator as region to be inpaint by using the ``black blurry hole'' as the prompt for Gounded-SAM~\cite{ren2024grounded} to find inpainting masks and therefore changes the background that should not be changed (shown in the third view). On the other hand, our 3DGIC locates the proper region to inpaint and produces smooth and high fidelity results.

\section{Limitations}
We now discuss the potential limitations of our 3DGIC. Since our 3DGIC uses the rendered depth map as guidance for the 3D inpainting process, the reliability of the rendered depth map becomes an important issue. As detailed in Sect.~\ref{subsec:backbone}, we combine the optimization technique introduced in Relightable Gaussians~\cite{gao2023relightable} to conduct a self-supervised loss for the predicted normal map and the rendered depth map to enhance the accuracy of the rendered depth map. However, if the input views are too sparse, the rendered depth map would not be guaranteed to be accurate, which hinders the inferring of Depth-Guided Mask and the achievement of cross-view consistency. Another potential limitation of our 3DGIC lies in the capability of the SAM~\cite{kirillov2023sam} model. As detailed in Sect.~\ref{subsec:backbone}, we use SAM to produce 2D segmentation masks and use these masks as supervision for our backbone 3DGS model so that we don't have to manually annotate the 2D object mask of the object to be removed like SPIn-NeRF~\cite{mirzaei2023spin}. However, if the object to be removed is too small, the SAM model would confuse it with other objects and not produce the correct segmentation mask for the object. To overcome the above limitations, studies on the production of reliable depth maps for 3DGS models with only sparse input views and producing a more accurate segmentation mask for any object would be possible directions to improve the quality of 3D Gaussian inpainting.

{
    \small
    \bibliographystyle{ieeenat_fullname}
    \bibliography{main}

\begin{thebibliography}{48}
\providecommand{\natexlab}[1]{#1}
\providecommand{\url}[1]{\texttt{#1}}
\expandafter\ifx\csname urlstyle\endcsname\relax
  \providecommand{\doi}[1]{doi: #1}\else
  \providecommand{\doi}{doi: \begingroup \urlstyle{rm}\Url}\fi

\bibitem[Barron et~al.(2021)Barron, Mildenhall, Tancik, Hedman, Martin-Brualla, and Srinivasan]{barron2021mip}
Jonathan~T Barron, Ben Mildenhall, Matthew Tancik, Peter Hedman, Ricardo Martin-Brualla, and Pratul~P Srinivasan.
\newblock Mip-nerf: A multiscale representation for anti-aliasing neural radiance fields.
\newblock In \emph{Proceedings of the IEEE/CVF International Conference on Computer Vision}, pages 5855--5864, 2021.

\bibitem[Barron et~al.(2022)Barron, Mildenhall, Verbin, Srinivasan, and Hedman]{barron2022mipnerf360}
Jonathan~T Barron, Ben Mildenhall, Dor Verbin, Pratul~P Srinivasan, and Peter Hedman.
\newblock Mip-nerf 360: Unbounded anti-aliased neural radiance fields.
\newblock In \emph{Proceedings of the IEEE Conference on Computer Vision and Pattern Recognition (CVPR)}, 2022.

\bibitem[Broll(2022)]{broll2022augmentedar2}
Wolfgang Broll.
\newblock Augmented reality.
\newblock In \emph{Virtual and Augmented Reality (VR/AR) Foundations and Methods of Extended Realities (XR)}, pages 291--329. Springer, 2022.

\bibitem[Burley and Studios(2012)]{burley2012brdf}
Brent Burley and Walt Disney~Animation Studios.
\newblock Physically-based shading at disney.
\newblock In \emph{Acm Siggraph}, 2012.

\bibitem[Chen et~al.(2022)Chen, Xu, Geiger, Yu, and Su]{chen2022tensorf}
Anpei Chen, Zexiang Xu, Andreas Geiger, Jingyi Yu, and Hao Su.
\newblock Tensorf: Tensorial radiance fields.
\newblock In \emph{European Conference on Computer Vision}, pages 333--350. Springer, 2022.

\bibitem[Chen and Wang(2024)]{chen2024surveygs}
Guikun Chen and Wenguan Wang.
\newblock A survey on 3d gaussian splatting.
\newblock \emph{arXiv preprint arXiv:2401.03890}, 2024.

\bibitem[Chen et~al.(2024)Chen, Loy, and Pan]{chen2024mvip}
Honghua Chen, Chen~Change Loy, and Xingang Pan.
\newblock Mvip-nerf: Multi-view 3d inpainting on nerf scenes via diffusion prior.
\newblock In \emph{Proceedings of the IEEE Conference on Computer Vision and Pattern Recognition (CVPR)}, 2024.

\bibitem[Corneanu et~al.(2024)Corneanu, Gadde, and Martinez]{corneanu2024latentpaint}
Ciprian Corneanu, Raghudeep Gadde, and Aleix~M Martinez.
\newblock Latentpaint: Image inpainting in latent space with diffusion models.
\newblock In \emph{Proceedings of the IEEE Winter Conference on Applications of Computer Vision (WACV)}, 2024.

\bibitem[Deng et~al.(2022)Deng, Liu, Zhu, and Ramanan]{deng2022depth}
Kangle Deng, Andrew Liu, Jun-Yan Zhu, and Deva Ramanan.
\newblock Depth-supervised nerf: Fewer views and faster training for free.
\newblock In \emph{Proceedings of the IEEE/CVF Conference on Computer Vision and Pattern Recognition}, pages 12882--12891, 2022.

\bibitem[Fridovich-Keil et~al.(2022)Fridovich-Keil, Yu, Tancik, Chen, Recht, and Kanazawa]{fridovich2022plenoxels}
Sara Fridovich-Keil, Alex Yu, Matthew Tancik, Qinhong Chen, Benjamin Recht, and Angjoo Kanazawa.
\newblock Plenoxels: Radiance fields without neural networks.
\newblock In \emph{Proceedings of the IEEE/CVF Conference on Computer Vision and Pattern Recognition}, pages 5501--5510, 2022.

\bibitem[Gao et~al.(2024)Gao, Gu, Lin, Zhu, Cao, Zhang, and Yao]{gao2023relightable}
Jian Gao, Chun Gu, Youtian Lin, Hao Zhu, Xun Cao, Li Zhang, and Yao Yao.
\newblock Relightable 3d gaussian: Real-time point cloud relighting with brdf decomposition and ray tracing.
\newblock \emph{Proceedings of the European Conference on Computer Vision (ECCV)}, 2024.

\bibitem[Hao et~al.(2021)Hao, Liu, Wu, Han, Chen, Chen, Chu, Tang, Yu, Chen, et~al.]{hao2021edgeflow}
Yuying Hao, Yi Liu, Zewu Wu, Lin Han, Yizhou Chen, Guowei Chen, Lutao Chu, Shiyu Tang, Zhiliang Yu, Zeyu Chen, et~al.
\newblock Edgeflow: Achieving practical interactive segmentation with edge-guided flow.
\newblock In \emph{Proceedings of the IEEE International Conference on Computer Vision (ICCV)}, 2021.

\bibitem[Heusel et~al.(2017)Heusel, Ramsauer, Unterthiner, Nessler, and Hochreiter]{heusel2017fid}
Martin Heusel, Hubert Ramsauer, Thomas Unterthiner, Bernhard Nessler, and Sepp Hochreiter.
\newblock Gans trained by a two time-scale update rule converge to a local nash equilibrium.
\newblock \emph{Advances in neural information processing systems}, 30, 2017.

\bibitem[Hu et~al.(2021)Hu, Shen, Wallis, Allen-Zhu, Li, Wang, Wang, and Chen]{hu2021lora}
Edward~J Hu, Yelong Shen, Phillip Wallis, Zeyuan Allen-Zhu, Yuanzhi Li, Shean Wang, Lu Wang, and Weizhu Chen.
\newblock Lora: Low-rank adaptation of large language models.
\newblock \emph{arXiv preprint arXiv:2106.09685}, 2021.

\bibitem[Kerbl et~al.(2023)Kerbl, Kopanas, Leimk{\"u}hler, and Drettakis]{kerbl202333dgs}
Bernhard Kerbl, Georgios Kopanas, Thomas Leimk{\"u}hler, and George Drettakis.
\newblock 3d gaussian splatting for real-time radiance field rendering.
\newblock \emph{ACM Transactions on Graphics (TOG)}, 2023.

\bibitem[Kerr et~al.(2023)Kerr, Kim, Goldberg, Kanazawa, and Tancik]{kerr2023lerf}
Justin Kerr, Chung~Min Kim, Ken Goldberg, Angjoo Kanazawa, and Matthew Tancik.
\newblock Lerf: Language embedded radiance fields.
\newblock In \emph{Proceedings of the IEEE International Conference on Computer Vision (ICCV)}, 2023.

\bibitem[Kirillov et~al.(2023)Kirillov, Mintun, Ravi, Mao, Rolland, Gustafson, Xiao, Whitehead, Berg, Lo, et~al.]{kirillov2023sam}
Alexander Kirillov, Eric Mintun, Nikhila Ravi, Hanzi Mao, Chloe Rolland, Laura Gustafson, Tete Xiao, Spencer Whitehead, Alexander~C Berg, Wan-Yen Lo, et~al.
\newblock Segment anything.
\newblock In \emph{Proceedings of the IEEE International Conference on Computer Vision (ICCV)}, 2023.

\bibitem[Lin et~al.(2024)Lin, Kim, Huang, Li, Ma, Kopf, Yang, and Tseng]{lin2024maldnerf}
Chieh~Hubert Lin, Changil Kim, Jia-Bin Huang, Qinbo Li, Chih-Yao Ma, Johannes Kopf, Ming-Hsuan Yang, and Hung-Yu Tseng.
\newblock Taming latent diffusion model for neural radiance field inpainting.
\newblock \emph{Proceedings of the European Conference on Computer Vision (ECCV)}, 2024.

\bibitem[Liu et~al.(2024)Liu, Ouyang, Wang, Cheng, Xiao, Zhu, Xue, Liu, Shen, and Cao]{liu2024infusion}
Zhiheng Liu, Hao Ouyang, Qiuyu Wang, Ka~Leong Cheng, Jie Xiao, Kai Zhu, Nan Xue, Yu Liu, Yujun Shen, and Yang Cao.
\newblock Infusion: Inpainting 3d gaussians via learning depth completion from diffusion prior.
\newblock \emph{arXiv preprint arXiv:2404.11613}, 2024.

\bibitem[Lugmayr et~al.(2022)Lugmayr, Danelljan, Romero, Yu, Timofte, and Van~Gool]{lugmayr2022repaint}
Andreas Lugmayr, Martin Danelljan, Andres Romero, Fisher Yu, Radu Timofte, and Luc Van~Gool.
\newblock Repaint: Inpainting using denoising diffusion probabilistic models.
\newblock In \emph{Proceedings of the IEEE Conference on Computer Vision and Pattern Recognition (CVPR)}, 2022.

\bibitem[Macedo and Apolinario(2021)]{macedo2021ar1}
M{\'a}rcio~CF Macedo and Antonio~L Apolinario.
\newblock Occlusion handling in augmented reality: past, present and future.
\newblock \emph{IEEE Transactions on Visualization and Computer Graphics}, 29\penalty0 (2):\penalty0 1590--1609, 2021.

\bibitem[Martin-Brualla et~al.(2021)Martin-Brualla, Radwan, Sajjadi, Barron, Dosovitskiy, and Duckworth]{martin2021nerf}
Ricardo Martin-Brualla, Noha Radwan, Mehdi~SM Sajjadi, Jonathan~T Barron, Alexey Dosovitskiy, and Daniel Duckworth.
\newblock Nerf in the wild: Neural radiance fields for unconstrained photo collections.
\newblock In \emph{Proceedings of the IEEE/CVF Conference on Computer Vision and Pattern Recognition}, pages 7210--7219, 2021.

\bibitem[Mildenhall et~al.(2021)Mildenhall, Srinivasan, Tancik, Barron, Ramamoorthi, and Ng]{mildenhall2021nerforiginal}
Ben Mildenhall, Pratul~P Srinivasan, Matthew Tancik, Jonathan~T Barron, Ravi Ramamoorthi, and Ren Ng.
\newblock Nerf: Representing scenes as neural radiance fields for view synthesis.
\newblock \emph{Communications of the ACM}, 65\penalty0 (1):\penalty0 99--106, 2021.

\bibitem[Mirzaei et~al.(2023{\natexlab{a}})Mirzaei, Aumentado-Armstrong, Brubaker, Kelly, Levinshtein, Derpanis, and Gilitschenski]{mirzaei2023referenceinpaint}
Ashkan Mirzaei, Tristan Aumentado-Armstrong, Marcus~A Brubaker, Jonathan Kelly, Alex Levinshtein, Konstantinos~G Derpanis, and Igor Gilitschenski.
\newblock Reference-guided controllable inpainting of neural radiance fields.
\newblock In \emph{Proceedings of the IEEE International Conference on Computer Vision (ICCV)}, 2023{\natexlab{a}}.

\bibitem[Mirzaei et~al.(2023{\natexlab{b}})Mirzaei, Aumentado-Armstrong, Derpanis, Kelly, Brubaker, Gilitschenski, and Levinshtein]{mirzaei2023spin}
Ashkan Mirzaei, Tristan Aumentado-Armstrong, Konstantinos~G Derpanis, Jonathan Kelly, Marcus~A Brubaker, Igor Gilitschenski, and Alex Levinshtein.
\newblock Spin-nerf: Multiview segmentation and perceptual inpainting with neural radiance fields.
\newblock In \emph{Proceedings of the IEEE Conference on Computer Vision and Pattern Recognition (CVPR)}, 2023{\natexlab{b}}.

\bibitem[Mirzaei et~al.(2024)Mirzaei, De~Lutio, Kim, Acuna, Kelly, Fidler, Gilitschenski, and Gojcic]{mirzaei2024reffusion}
Ashkan Mirzaei, Riccardo De~Lutio, Seung~Wook Kim, David Acuna, Jonathan Kelly, Sanja Fidler, Igor Gilitschenski, and Zan Gojcic.
\newblock Reffusion: Reference adapted diffusion models for 3d scene inpainting.
\newblock \emph{arXiv preprint arXiv:2404.10765}, 2024.

\bibitem[M{\"u}ller et~al.(2022)M{\"u}ller, Evans, Schied, and Keller]{muller2022instant}
Thomas M{\"u}ller, Alex Evans, Christoph Schied, and Alexander Keller.
\newblock Instant neural graphics primitives with a multiresolution hash encoding.
\newblock \emph{ACM Transactions on Graphics (ToG)}, 41\penalty0 (4):\penalty0 1--15, 2022.

\bibitem[Paszke et~al.(2019)Paszke, Gross, Massa, Lerer, Bradbury, Chanan, Killeen, Lin, Gimelshein, Antiga, et~al.]{paszke2019pytorch}
Adam Paszke, Sam Gross, Francisco Massa, Adam Lerer, James Bradbury, Gregory Chanan, Trevor Killeen, Zeming Lin, Natalia Gimelshein, Luca Antiga, et~al.
\newblock Pytorch: An imperative style, high-performance deep learning library.
\newblock \emph{Advances in neural information processing systems}, 32, 2019.

\bibitem[Podell et~al.(2024)Podell, English, Lacey, Blattmann, Dockhorn, M{\"u}ller, Penna, and Rombach]{podell2023sdxl}
Dustin Podell, Zion English, Kyle Lacey, Andreas Blattmann, Tim Dockhorn, Jonas M{\"u}ller, Joe Penna, and Robin Rombach.
\newblock Sdxl: Improving latent diffusion models for high-resolution image synthesis.
\newblock \emph{Proceedings of the International Conference on Learning Representations (ICLR)}, 2024.

\bibitem[Qin et~al.(2024)Qin, Li, Zhou, Wang, and Pfister]{qin2024langsplat}
Minghan Qin, Wanhua Li, Jiawei Zhou, Haoqian Wang, and Hanspeter Pfister.
\newblock Langsplat: 3d language gaussian splatting.
\newblock In \emph{Proceedings of the IEEE Conference on Computer Vision and Pattern Recognition (CVPR)}, 2024.

\bibitem[Reiser et~al.(2021)Reiser, Peng, Liao, and Geiger]{reiser2021kilonerf}
Christian Reiser, Songyou Peng, Yiyi Liao, and Andreas Geiger.
\newblock Kilonerf: Speeding up neural radiance fields with thousands of tiny mlps.
\newblock In \emph{Proceedings of the IEEE/CVF International Conference on Computer Vision}, pages 14335--14345, 2021.

\bibitem[Ren et~al.(2024)Ren, Liu, Zeng, Lin, Li, Cao, Chen, Huang, Chen, Yan, et~al.]{ren2024grounded}
Tianhe Ren, Shilong Liu, Ailing Zeng, Jing Lin, Kunchang Li, He Cao, Jiayu Chen, Xinyu Huang, Yukang Chen, Feng Yan, et~al.
\newblock Grounded sam: Assembling open-world models for diverse visual tasks.
\newblock \emph{arXiv preprint arXiv:2401.14159}, 2024.

\bibitem[Rombach et~al.(2022)Rombach, Blattmann, Lorenz, Esser, and Ommer]{ldm}
Robin Rombach, Andreas Blattmann, Dominik Lorenz, Patrick Esser, and Bj{\"o}rn Ommer.
\newblock High-resolution image synthesis with latent diffusion models.
\newblock In \emph{Proceedings of the IEEE Conference on Computer Vision and Pattern Recognition (CVPR)}, 2022.

\bibitem[Song et~al.(2021)Song, Meng, and Ermon]{song2020ddim}
Jiaming Song, Chenlin Meng, and Stefano Ermon.
\newblock Denoising diffusion implicit models.
\newblock \emph{Proceedings of the International Conference on Learning Representations (ICLR)}, 2021.

\bibitem[Sun et~al.(2022)Sun, Sun, and Chen]{sun2022direct}
Cheng Sun, Min Sun, and Hwann-Tzong Chen.
\newblock Direct voxel grid optimization: Super-fast convergence for radiance fields reconstruction.
\newblock In \emph{Proceedings of the IEEE/CVF Conference on Computer Vision and Pattern Recognition}, pages 5459--5469, 2022.

\bibitem[Suvorov et~al.(2022)Suvorov, Logacheva, Mashikhin, Remizova, Ashukha, Silvestrov, Kong, Goka, Park, and Lempitsky]{lama}
Roman Suvorov, Elizaveta Logacheva, Anton Mashikhin, Anastasia Remizova, Arsenii Ashukha, Aleksei Silvestrov, Naejin Kong, Harshith Goka, Kiwoong Park, and Victor Lempitsky.
\newblock Resolution-robust large mask inpainting with fourier convolutions.
\newblock In \emph{Proceedings of the IEEE Winter Conference on Applications of Computer Vision (WACV)}, 2022.

\bibitem[Wang et~al.(2024{\natexlab{a}})Wang, Zhang, Abboud, and S{\"u}sstrunk]{wang2024innerf360}
Dongqing Wang, Tong Zhang, Alaa Abboud, and Sabine S{\"u}sstrunk.
\newblock Innerf360: Text-guided 3d-consistent object inpainting on 360-degree neural radiance fields.
\newblock In \emph{Proceedings of the IEEE Conference on Computer Vision and Pattern Recognition (CVPR)}, 2024{\natexlab{a}}.

\bibitem[Wang et~al.(2024{\natexlab{b}})Wang, Wu, Zhang, and Xu]{wang2024gscream}
Yuxin Wang, Qianyi Wu, Guofeng Zhang, and Dan Xu.
\newblock Gscream: Learning 3d geometry and feature consistent gaussian splatting for object removal.
\newblock \emph{Proceedings of the European Conference on Computer Vision (ECCV)}, 2024{\natexlab{b}}.

\bibitem[Weber et~al.(2024)Weber, Holynski, Jampani, Saxena, Snavely, Kar, and Kanazawa]{weber2024nerfiller}
Ethan Weber, Aleksander Holynski, Varun Jampani, Saurabh Saxena, Noah Snavely, Abhishek Kar, and Angjoo Kanazawa.
\newblock Nerfiller: Completing scenes via generative 3d inpainting.
\newblock In \emph{Proceedings of the IEEE Conference on Computer Vision and Pattern Recognition (CVPR)}, pages 20731--20741, 2024.

\bibitem[Weder et~al.(2023)Weder, Garcia-Hernando, Monszpart, Pollefeys, Brostow, Firman, and Vicente]{weder2023removingnerf}
Silvan Weder, Guillermo Garcia-Hernando, Aron Monszpart, Marc Pollefeys, Gabriel~J Brostow, Michael Firman, and Sara Vicente.
\newblock Removing objects from neural radiance fields.
\newblock In \emph{Proceedings of the IEEE Conference on Computer Vision and Pattern Recognition (CVPR)}, 2023.

\bibitem[Xie et~al.(2023)Xie, Zhang, Lin, Hinz, and Zhang]{xie2023smartbrush}
Shaoan Xie, Zhifei Zhang, Zhe Lin, Tobias Hinz, and Kun Zhang.
\newblock Smartbrush: Text and shape guided object inpainting with diffusion model.
\newblock In \emph{Proceedings of the IEEE Conference on Computer Vision and Pattern Recognition (CVPR)}, 2023.

\bibitem[Yang et~al.(2023{\natexlab{a}})Yang, Gu, Zhang, Zhang, Chen, Sun, Chen, and Wen]{yang2023paintbyexample}
Binxin Yang, Shuyang Gu, Bo Zhang, Ting Zhang, Xuejin Chen, Xiaoyan Sun, Dong Chen, and Fang Wen.
\newblock Paint by example: Exemplar-based image editing with diffusion models.
\newblock In \emph{Proceedings of the IEEE Conference on Computer Vision and Pattern Recognition (CVPR)}, 2023{\natexlab{a}}.

\bibitem[Yang et~al.(2023{\natexlab{b}})Yang, Chen, and Liao]{yang2023unipaint}
Shiyuan Yang, Xiaodong Chen, and Jing Liao.
\newblock Uni-paint: A unified framework for multimodal image inpainting with pretrained diffusion model.
\newblock In \emph{Proceedings of the 31st ACM International Conference on Multimedia}, 2023{\natexlab{b}}.

\bibitem[Ye et~al.(2024)Ye, Danelljan, Yu, and Ke]{ye2023gaussiangrouping}
Mingqiao Ye, Martin Danelljan, Fisher Yu, and Lei Ke.
\newblock Gaussian grouping: Segment and edit anything in 3d scenes.
\newblock \emph{Proceedings of the European Conference on Computer Vision (ECCV)}, 2024.

\bibitem[Yin et~al.(2023)Yin, Fu, Yang, and Lin]{yin2023ornerf}
Youtan Yin, Zhoujie Fu, Fan Yang, and Guosheng Lin.
\newblock Or-nerf: Object removing from 3d scenes guided by multiview segmentation with neural radiance fields.
\newblock \emph{arXiv preprint arXiv:2305.10503}, 2023.

\bibitem[Yu et~al.(2021)Yu, Li, Tancik, Li, Ng, and Kanazawa]{yu2021plenoctrees}
Alex Yu, Ruilong Li, Matthew Tancik, Hao Li, Ren Ng, and Angjoo Kanazawa.
\newblock Plenoctrees for real-time rendering of neural radiance fields.
\newblock In \emph{Proceedings of the IEEE/CVF International Conference on Computer Vision}, pages 5752--5761, 2021.

\bibitem[Yu et~al.(2024)Yu, Chen, Huang, Sattler, and Geiger]{yu2024mipsplat}
Zehao Yu, Anpei Chen, Binbin Huang, Torsten Sattler, and Andreas Geiger.
\newblock Mip-splatting: Alias-free 3d gaussian splatting.
\newblock In \emph{Proceedings of the IEEE Conference on Computer Vision and Pattern Recognition (CVPR)}, 2024.

\bibitem[Zhang et~al.(2018)Zhang, Isola, Efros, Shechtman, and Wang]{lpips}
Richard Zhang, Phillip Isola, Alexei~A Efros, Eli Shechtman, and Oliver Wang.
\newblock The unreasonable effectiveness of deep features as a perceptual metric.
\newblock In \emph{Proceedings of the IEEE Conference on Computer Vision and Pattern Recognition (CVPR)}, 2018.

\end{thebibliography}
}

\end{document}